\pdfoutput=1

\documentclass[11pt]{article}

\usepackage{emnlp2023}

\usepackage{times}
\usepackage{latexsym}
\usepackage{amsmath}
\usepackage{bm}
\usepackage{times}
\usepackage{latexsym}
\usepackage{amsmath}
\usepackage{graphicx}
\usepackage{bm}
\usepackage{amsfonts,amssymb}
\usepackage{bbm}
\usepackage{booktabs}
\usepackage{booktabs,siunitx}
\usepackage{multirow}
\usepackage{subfigure}
\usepackage{parskip}
\usepackage{graphicx}
\usepackage{subfigure}
\usepackage{graphicx}
\usepackage[T1]{fontenc}
\usepackage[utf8]{inputenc}
\DeclareUnicodeCharacter{0307}{.} 
\usepackage{microtype}
\usepackage{array}
\usepackage{booktabs}
\usepackage{multirow}
\usepackage{algorithm}
\usepackage{enumitem}
\usepackage{algorithmic}
\usepackage{booktabs}
\usepackage{colortbl}
\usepackage{siunitx}
\PassOptionsToPackage{table}{xcolor}
\usepackage{xcolor}

\usepackage[T1]{fontenc}

\usepackage[utf8]{inputenc}

\usepackage{microtype}

\usepackage{inconsolata}

\title{

ProMedTS: A Self-Supervised, Prompt-Guided Multimodal Approach for Integrating Medical Text and Time Series
}

\author{Shuai Niu$^1$, Jing Ma$^1$, Hongzhan Lin$^1$, 
    Liang Bai$^2$, Zhihua Wang$^3$, \\ \textbf{Wei Bi$^4$, Yida Xu$^1$,} \textbf{Guo Li$^5$,} and \textbf{Xian Yang$^6$}\thanks{* This is the corresponding author.}
    \\
    Hong Kong Baptist University$^1$,
    Shanxi University$^2$ ,\\
    Shanghai Institute for Advanced Study of Zhejiang University$^3$,Tencent AI Lab$^4$\\
    Manchester Metropolitan University$^5$, The University of Manchester$^6$\\
    \texttt{\{cssniu,majing\}@comp.hkbu.edu.hk, xian.yang@manchester.ac.uk}\\
}

\begin{document}
\maketitle
\begin{abstract} Large language models (LLMs) have shown remarkable performance in vision-language tasks, but their application in the medical field remains underexplored, particularly for integrating structured time series data with unstructured clinical notes. In clinical practice, dynamic time series data, such as lab test results, capture critical temporal patterns, while clinical notes provide rich semantic context. Merging these modalities is challenging due to the inherent differences between continuous signals and discrete text. To bridge this gap, we introduce ProMedTS, a novel self-supervised multimodal framework that employs prompt-guided learning to unify these heterogeneous data types. Our approach leverages lightweight anomaly detection to generate anomaly captions that serve as prompts, guiding the encoding of raw time series data into informative prompt embeddings. These prompt embeddings are aligned with textual representations in a shared latent space, preserving fine-grained temporal nuances alongside semantic insights. Furthermore, our framework incorporates tailored self-supervised objectives to enhance both intra- and inter-modal alignment. We evaluate ProMedTS on disease diagnosis tasks using real-world datasets, and the results demonstrate that our method consistently outperforms state-of-the-art approaches. \end{abstract}

\begin{figure}[t] \centering \includegraphics[scale=0.23]{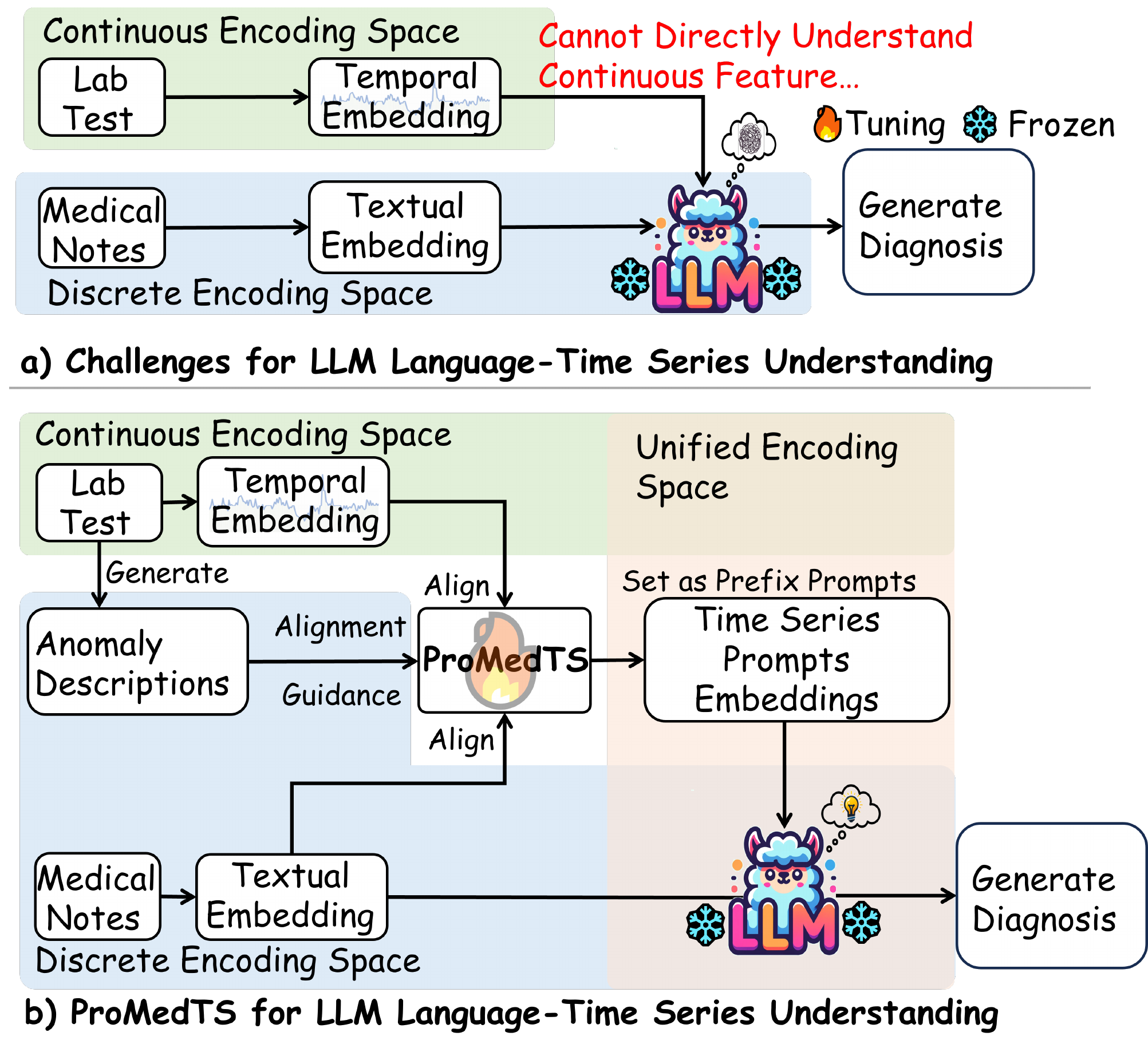} \caption{(a) LLMs struggle to process continuous time series data due to modality gaps with discrete textual representations. (b) ProMedTS bridges this gap by leveraging anomaly descriptions and time series prompts, aligning structured EHR data with clinical notes for improved multimodal understanding. } \label{Intro} \end{figure}

\section{Introduction}

Recent advancements in natural language processing (NLP) have revolutionized healthcare by enabling deeper insights into electronic health records (EHRs). EHRs combine structured data, such as time series laboratory (lab) test results, with unstructured data, including clinical notes and medical images. While large language models (LLMs) excel at processing unstructured text \cite{nori2023capabilities,singhal2023large} and vision transformers have driven progress in medical image analysis \cite{wang2022medclip,chen2021transunet}, integrating time series data with text remains a challenge. Unlike text, which is composed of discrete tokens, time series data contains continuous signals with temporal dependencies as illustrated in Figure~\ref{Intro}(a).

Current multimodal learning approaches, especially contrastive learning methods \cite{radford2021learning,li2023blip2}, have been effective in aligning vision data and text data. However, they are less suited to bridge the gap between time series data and text data. Time series data require fine-grained temporal representations in a high-dimensional space and are often irregularly sampled, exhibit diverse frequencies, and include missing values \cite{harutyunyan2019multitask}. In addition, the lack of large-scale paired datasets that link raw time series with textual descriptions further hampers LLMs from incorporating structured information into clinical decision-making \cite{niu2024ehr}. Without an effective fusion mechanism, LLMs cannot fully exploit the rich temporal patterns in structured EHR data.

To address these challenges, we propose ProMedTS, a self-supervised and prompt-guided framework designed to unify medical notes and time series lab tests for natural comprehension by LLMs. As illustrated in Figure~\ref{Intro}(b), rather than only feeding raw time series data directly into LLMs, our framework introduces \emph{anomaly descriptions} as a modality bridge to facilitate alignment between text and time series data and to aid in generating time series prompt embeddings.  These descriptions are produced using anomaly detection technology \cite{vinutha2018detection}, which converts continuous signals into human-readable summaries that encompass coarse-grained time series patterns. The process consists of two steps. First, anomaly descriptions establish a direct connection between time series lab tests and medical notes. Second, the generated time series prompt embeddings contain both coarse-grained anomaly information and fine-grained time series variation patterns. These embeddings are then appended as prefix tokens to the input of the LLM. This approach integrates structured time series information into the language modeling process without modifying the LLM architecture, thereby unifying both modalities within the same encoding space and enhancing clinical decision-making.

We optimize ProMedTS with three self-supervised learning objectives for learning the time series prompt embeddings. A contrastive loss maps textual and time series modalities into a shared latent space to learn the coarse-grained lab test information. An anomaly-time series matching loss links numeric lab tests with their corresponding anomaly descriptions to reinforce consistency and learn the fine-grained lab test information. {Finally, an anomaly caption generation loss enhances the representation of multi-granularity time series information within the time series prompt embeddings.} Together, these objectives enable LLMs to process both structured and unstructured EHR data more effectively, addressing the gap between language and time series representations in healthcare applications.

\begin{itemize}[itemsep=2pt,topsep=0pt,parsep=0pt] \item We propose ProMedTS, a self-supervised framework that integrates structured time series and unstructured textual EHR data into LLMs without changing their architectures. \item We introduce anomaly descriptions as a textual bridge to align time series data with clinical notes, supported by three self-supervised objectives. \item We demonstrate that ProMedTS significantly improves disease diagnosis on MIMIC-III and MIMIC-IV, setting a new benchmark for multimodal EHR learning. \end{itemize}

\section{Related Work}
\subsection{Multimodal Learning in Healthcare}
The increasing diversity of EHR data has led to significant advancements in multimodal learning for healthcare applications. MedCLIP \cite{wang2022medclip} employs semantic contrastive learning to align medical images with textual reports, while RAIM \cite{qiao2019mnn} and GLoRIA \cite{huang2021gloria} integrate numerical or image data with text using attention mechanisms. LDAM \cite{niu2021label} further extends these approaches by leveraging cross-attention with disease labels to fuse features from lab tests and clinical notes. EHR-KnowGen \cite{niu2024ehr} transforms structured lab data into text and incorporates external knowledge for improved modality fusion. Despite these advancements, achieving a unified latent embedding that effectively captures interactions across diverse modalities remains a key challenge in multimodal EHR processing.
\subsection{ Healthcare with LLMs}
Beyond multimodal learning, recent research has explored generative approaches to healthcare modeling. Conventional methods have primarily relied on discriminative models for disease risk assessment and diagnosis \cite{choi2016retain, niu2024enhancing,qiao2019mnn}. However, generative models are increasingly being adopted, as demonstrated by Clinical CoT \cite{kwon2024large}, applying LLMs for disease diagnosis generation. Reinforcement learning from human feedback (RLHF) \cite{ouyang2022training} and Chain-of-Thought (CoT) prompting \cite{wei2022chain} have further enhanced medical reasoning capabilities in models such as GatorTron \cite{yang2022large}, MedPalm \cite{singhal2023large}, and GPT4-Med \cite{nori2023capabilities}. While these models excel in medical question-answering, they remain limited in real-world direct disease diagnosis and multimodal EHR processing. EHR-KnowGen \cite{niu2024ehr} reframes disease diagnosis as a text-to-text generation problem but overlooks the crucial temporal details embedded in time series lab tests, underscoring the need for more effective and dedicated multimodal fusion strategies.

\begin{figure*}[t]
\centering
\centerline{\includegraphics[scale=0.233]{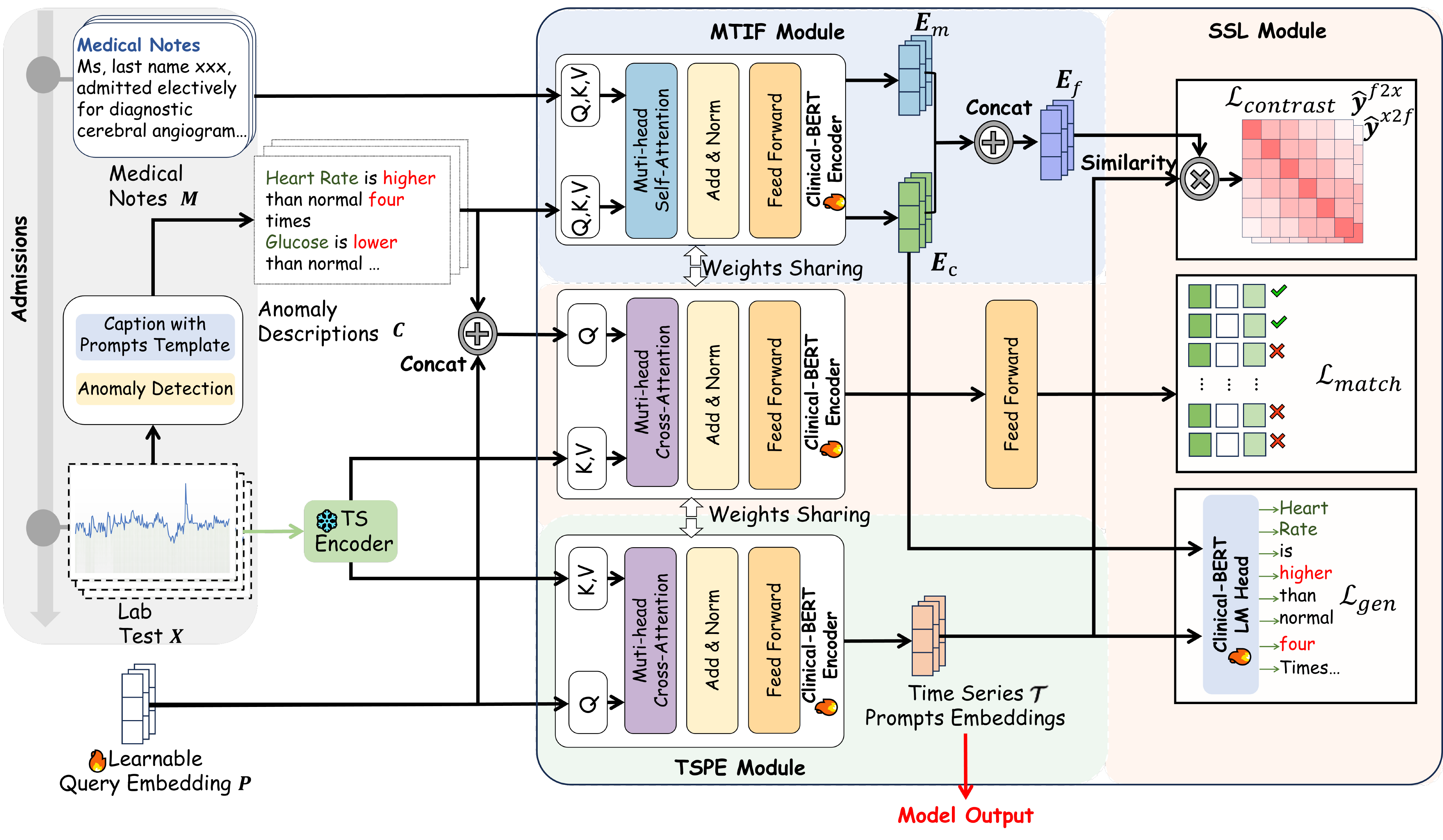}}
\caption{
The ProMedTS model comprises three modules: the Time Series Prompt Embedding (TSPE) module, the Multimodal Textual Information Fusion (MTIF) module, and the Self-supervised Learning (SSL) module. The MTIF module utilizes Clinical-BERT to encode medical notes $\bm{M}$, lab test data $\bm{X}$, and anomaly descriptions $\bm{C}$ to generate time series prompt embeddings $\bm{\mathcal{T}}$. 
}
\label{model1}
\end{figure*}

\section{Methodology}
In this section, we present the ProMedTS framework for unifying heterogeneous EHR data through prompt-guided learning. We begin by defining the problem and describing the model inputs, then provide a high-level overview of the architecture. In subsequent sections, we detail each module and discuss how these components are applied to downstream tasks such as disease diagnosis.
\subsection{Problem Definition}
We introduce ProMedTS, aiming to reduce discrepancies between language and time series EHRs. Specifically, it leverages anomaly captions and generates time series prompt embeddings to unify both modalities in a shared latent space. The inputs to ProMedTS, denoted by $\{\bm{M}, \bm{X}\}$, include medical notes $\bm{M} \in \mathbb{R}^{B \times N_m}$ (where $B$ is the batch size and $N_m$ is the number of tokens) and numeric lab test data $\bm{X} \in \mathbb{R}^{B \times L \times N_x}$ (where $L$ is the sequence length and $N_x$ is the number of lab test variants). Additionally, a lightweight anomaly detection \cite{vinutha2018detection} is employed to generate textual descriptions of anomalies $\bm{C} \in \mathbb{R}^{B \times N_c}$ (details in Appendix A.2). ProMedTS also uses learnable time series query embeddings $\bm{P} \in \mathbb{R}^{B \times N_p \times D}$, which are transformed into time series prompt embeddings $\bm{\mathcal{T}} \in \mathbb{R}^{B \times N_p \times D}$, where $N_p$ is the query length and $D$ is the hidden dimension.

\subsection{Model Overview}
Figure~\ref{model1} illustrates the overview of ProMedTS, which comprises three main modules.  {Three modules share the same Clinical-BERT\cite{alsentzer2019publicly} structured model for clinical tokens embedding and are extended to support cross-attention, self-attention, and prompt generation.} The Time Series Prompt Embedding (TSPE) module applies a cross-attention mechanism to convert raw lab test data into prompt embeddings, preserving key temporal features. The Multimodal Textual Information Fusion (MTIF) module encodes and merges medical notes with anomaly captions in a unified latent space, facilitating the extraction of complementary semantic information. Finally, the Self-supervised Learning (SSL) module employs tailored loss functions to bridge the modality gap and maintain multi-granularity temporal details in the learned representations. These modules work in tandem to achieve robust alignment and fusion of heterogeneous EHRs, and the following sections provide in-depth explanations of each component and its applications.

\subsection{ Time Series Prompt Embedding}

The objective of the TSPE module is to extract and encapsulate the inherent fine-grained temporal information from time series lab tests into time series prompt embeddings. Let $\{\bm{X}, \bm{P}\}$ represent the module inputs. The numeric lab test data $\bm{X}$ is first processed by a time series encoder (TSE) using PatchTST \cite{nie2022time}. {In parallel, the learnable query embeddings $\bm{P}$, initialized using vectors slicing from the Clinical-BERT word embedding layer, serve as query tokens in the cross-attention mechanism, guiding the selection of relevant temporal features by attending to time series lab tests encoded by TSE. To generate the final prompt embedding $\bm{\mathcal{T}}$, we extend the multi-head self-attention encoder of Clinical-BERT to support a multi-head cross-attention mechanism, following a strategy similar to that adopted in \cite{li2023blip2}. We designate $\bm{X}$ as both key and value while $\bm{P}$ serves as the query:}
\begin{equation}
\bm{\mathcal{T}} = \text{Clinical-BERT}\bigl(\bm{P}, \text{TSE}(\bm{X}), \text{TSE}(\bm{X})\bigr).
\label{eq1}
\end{equation}
This design ensures that the rich temporal patterns in $\bm{X}$ are captured within $\bm{\mathcal{T}}$, enabling subsequent modules to leverage these features effectively.

\subsection{Multimodal Textual Information Fusion }
 The MTIF module is designed to fuse medical notes and anomaly descriptions effectively. \textcolor{black}{We use the anomaly captioning method to generate anomaly descriptions, as illustrated in Figure~\ref{model1}.} The inputs to the MTIF module are medical notes $\bm{M}$ and lab test anomaly descriptions $\bm{C}$, which are encoded separately by Clinical-BERT via the multi-head self-attention mechanism:
\begin{equation}
\begin{aligned}
&\bm{E}_m = \text{Clinical-BERT}(\bm{M},\bm{M},\bm{M}), \\
&\bm{E}_c = \text{Clinical-BERT}(\bm{C},\bm{C},\bm{C}), 
\label{eq2}
\end{aligned}
\end{equation}
where $\bm{E}_m \in \mathbb{R}^{B \times N_m \times D}$ and $\bm{E}_c \in \mathbb{R}^{B \times N_c \times D}$. The repeated inputs indicate that the key, query, and value matrices are identical for the self-attention mechanism.
This structure enables the model to encode each type of textual information independently while capturing the inherent characteristics and context of each input. The combined textual representation is then derived from $\bm{E}_m$ and $\bm{E}_c$:
\begin{equation}
\bm{E}_f = AVG([\bm{E}_m \oplus \bm{E}_c]),
\label{eq3}
\end{equation}
where $\bm{E}_f \in \mathbb{R}^{B \times D}$, with $\oplus$ indicating concatenation, and $AVG$ representing average pooling.

\subsection{Self-Supervised Learning}

This module addresses the modality gap between textual and time series EHR data using three specialized loss functions. By simultaneously aligning cross-modal representations and preserving coarse- and fine-grained temporal details, the model learns to capture both semantic and temporal nuances.

\subsubsection{Cross-Modal Contrastive Alignment}

To promote cross-modal alignment and preserve coarse-grained temporal information into time series prompt embeddings, we design a contrastive loss that brings multimodal textual embedding and time series embeddings closer when they originate from the same patient and pushes them apart otherwise. We first compute similarity matrices by multiplying the fused text representation $\bm{E}_f$ with the time series prompt embeddings $\bm{\mathcal{T}}$:
\begin{equation}
\begin{aligned}
\bm{g}_{(\bm{E}_f,\bm{X})} &= \max\Bigl(\bigl[\bm{E}_f\,\bm{\mathcal{T}}^{(1)T},\ldots,\bm{E}_f\,\bm{\mathcal{T}}^{(N_p)T}\bigr]\Bigr),\\
\bm{g}_{(\bm{X},\bm{E}_f)} &= \max\Bigl(\bigl[\bm{\mathcal{T}}^{(1)}\,\bm{E}_f^T,\ldots,\bm{\mathcal{T}}^{(N_p)}\,\bm{E}_f^T\bigr]\Bigr),
\end{aligned}
\label{eq4}
\end{equation}
where the $\max$ operator performs max-pooling across $N_p$ dimensions, yielding $\bm{g}_{(\bm{E}_f,\bm{X})}$ and $\bm{g}_{(\bm{X},\bm{E}_f)} \in \mathbb{R}^{B \times B}$. Note that $\bm{g}_{(\bm{E}_f,\bm{X})}$ measures text-to-time series similarity (by fixing $\bm{E}_f$ and iterating over $\bm{\mathcal{T}}$), while $\bm{g}_{(\bm{X},\bm{E}_f)}$ captures time-series-to-text similarity (by fixing $\bm{\mathcal{T}}$ and iterating over $\bm{E}_f$). This process is the same as that used in vision-language contrastive learning \cite{radford2021learning, li2023blip2}. We then apply the SoftMax function to generate two distinct sets of logits:
\begin{equation}
\begin{aligned}
\hat{\bm{y}}_c^{f2x} &= \text{SoftMax}\bigl(\bm{g}_{(\bm{E}_f,\bm{X})}\bigr),\\
\hat{\bm{y}}_c^{x2f} &= \text{SoftMax}\bigl(\bm{g}_{(\bm{X},\bm{E}_f)}\bigr).
\end{aligned}
\label{eq5}
\end{equation}
{ Let $\bm{y}_c^{f2x}$ and $\bm{y}_c^{x2f}$denote the ground truth labels indicating whether the pairs correspond to the same patient in a training batch (1 if matched, 0 otherwise). We use cross-entropy $H(\cdot)$ to define the contrastive loss:}
\begin{equation}
\begin{aligned}
\mathcal{L}_{contrast} \;&=\; \tfrac{1}{2}\,\mathbb{E}\Bigl[
  H\bigl(\bm{y}_c^{f2x},\,\hat{\bm{y}}_c^{f2x}\bigr)\;\\
  &+\;
  H\bigl(\bm{y}_c^{x2f},\,\hat{\bm{y}}_c^{x2f}\bigr)
\Bigr].
\end{aligned}
\label{eq6}
\end{equation}

\subsubsection{Intra-Modal Matching}
To further capture intra-modality consistency and enhance alignment with fine-grained temporal information, we align lab tests with their corresponding anomaly descriptions and time series prompt embeddings. This alignment is modeled as a binary classification task, distinguishing matched from unmatched pairs of lab tests and anomaly captions. { Following \citet{li2021align}, we employ a negative mining strategy to generate labels $\bm{y}_m$ by selecting the most similar pairs in a training batch as negative samples, where the top 1-ranked pair is labeled as 1 and the others as 0, based on the similarity computed in Equation \ref{eq4}.} We employ Clinical-BERT’s cross-attention, where the concatenation of $\bm{C}$ and $\bm{P}$ serves as the query (aiming to align the time series prompt embedding more closely with the textual encoding space during training), and the encoded time series $\bm{X}$ is used as both key and value. A Multilayer Perceptron (MLP) classifier with softmax activation, denoted $f_{match}$, predicts the probability $\hat{\bm{y}}_m$:
\begin{equation}
\begin{aligned}
\hat{\bm{y}}_m =
f_{match}\Bigl(\text{Clinical-BERT}\bigl(
  f_\mathcal{W}(\bm{C}) \oplus \bm{P}, \\
  \text{TSE}(\bm{X}),
  \text{TSE}(\bm{X})
\bigr)\Bigr),
\end{aligned}
\label{eq7}
\end{equation}
where $f_\mathcal{W}$ is the word embedding layer in Clinical-BERT. We define the matching loss as:
\begin{equation}
\mathcal{L}_{match} \;=\;
\mathbb{E}\bigl[H\bigl(\bm{y}_m,\,\hat{\bm{y}}_m\bigr)\bigr],
\label{eq8}
\end{equation}
where $\bm{y}_m$ is the one-hot ground truth label.

\subsubsection{Anomaly Description Reconstruction}
To ensure the time series prompt embeddings encode both coarse anomaly descriptions and fine-grained temporal details, we reconstruct anomaly captions from the learned embeddings. This step helps unify language tokens and time series representations in a shared space. Specifically, we use Clinical-BERT with a language model head $f_{head}$, setting $\bm{E}_c$ as the query and $\bm{\mathcal{T}}$ as key and value:
\begin{equation}
\begin{small}
\mathcal{L}_{gen} =
\mathbb{E}\Bigl[
  H\bigl(\bm{C},
    f_{head}\bigl(\text{Clinical-BERT}(\bm{E}_c,
      \bm{\mathcal{T}},\bm{\mathcal{T}})\bigr)
  \bigr)
\Bigr].
\end{small}
\label{eq9}
\end{equation}
This objective is a standard language model generation loss, computed as cross-entropy between the predicted token distribution and the ground truth tokens, encouraging the model to generate accurate textual descriptions, thereby reinforcing alignment between time series prompts and language tokens.

\textbf{Overall Loss:}
We combine these objectives into a single training loss:
\begin{equation}
\mathcal{L}_{total}=
\alpha\,\mathcal{L}_{contrast}
\;+\;\beta\,\mathcal{L}_{match}
\;+\;\gamma\,\mathcal{L}_{gen},
\label{eq10}
\end{equation}
where $\alpha$, $\beta$, and $\gamma$ are hyperparameters balancing the three losses (see Appendix A.6). Our training algorithm aims to minimize $\mathcal{L}_{total}$ across all samples (details in Appendix A.1).

\subsection{LLM-based Disease Diagnosis with ProMedTS}
\begin{figure}[t]
\centering
\centerline{\includegraphics[scale=0.24]{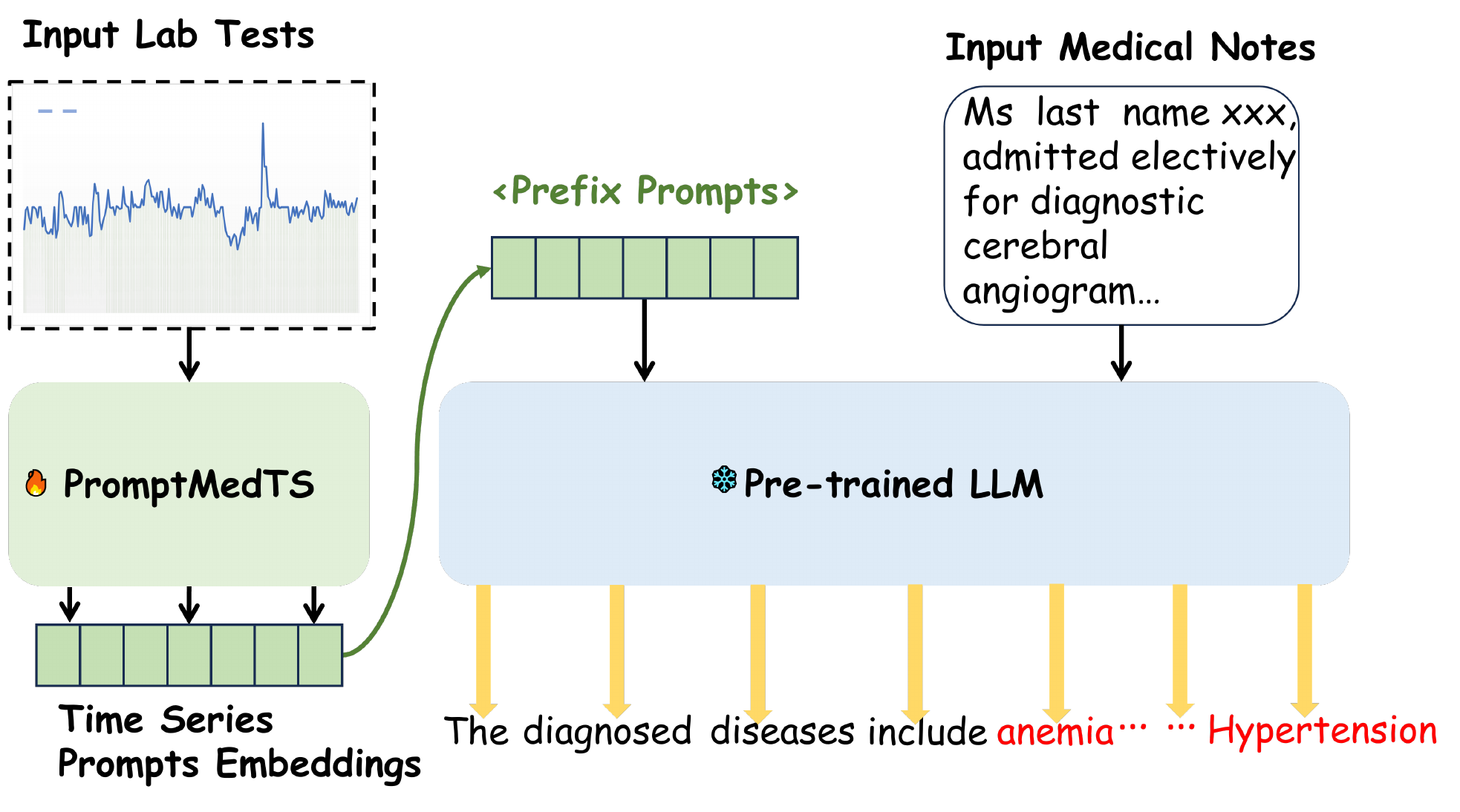}}
\caption{ProMedTS for empowering LLMs to in disease diagnosis.}
\label{model2}
\end{figure}

To illustrate the practical effectiveness of ProMedTS in unifying textual and time series data, we employ a pre-trained, frozen LLM model for disease diagnosis. As depicted in Figure~\ref{model2}, during ProMedTS's fine-tuning, it first transforms numeric lab test results into time series prompt embeddings, which are subsequently aligned with the LLM's input dimensions through a tunable fully connected layer. These embeddings then serve as prefix soft prompts, concatenated with the medical notes so that the model can ingest structured signals from time series alongside unstructured clinical text. By bridging language and time series modalities, the LLM can process both inputs concurrently, leveraging complementary information for enhanced diagnostic accuracy.

\begin{table*}[t]
\small
\setlength\tabcolsep{1.6pt}
\renewcommand\arraystretch{1}
\centering

\begin{tabular}{l|c|cc|cc|ccc|ccc}
\toprule[1pt]

\multirow{2}{*}{\textbf{Models}} & \multirow{2}{*}{\textbf{Size}} &\multicolumn{2}{c|}{\textbf{Type}} &\multicolumn{2}{c|}{\textbf{Modality}} &\multicolumn{3}{c|}{\textbf{Micro}} &\multicolumn{3}{c}{\textbf{Macro}}  \\
                            \cline{3-12}
                          && \multicolumn{1}{c|}{\textbf{CLS}} & \textbf{GEN} & \multicolumn{1}{c|}{\textbf{Lab}} &\multicolumn{1}{c|}{\textbf{Note}} & \multicolumn{1}{c}{\textbf{Precision}} & \multicolumn{1}{c}{\textbf{Recall}}   & \multicolumn{1}{c|}{\textbf{F1}} & \multicolumn{1}{c}{\textbf{Precision}} & \multicolumn{1}{c}{\textbf{Recall}}   & \multicolumn{1}{c}{\textbf{F1}}   \\ 
                         \cline{1-12}
 \multicolumn{12}{c}{\textbf{ MIMIC-III}}  \\ 
                         \cline{1-12}
\textbf{GRU}  &  7.9M  & \checkmark &  & \checkmark &  & 46.41$_{(3.48)}$ & 21.88$_{(3.59)}$ & 29.43$_{(1.89)}$ & 30.47$_{(4.23)}$ & 13.00$_{(1.14)}$ & 14.59$_{(1.48)}$  \\
\textbf{PatchTST}  &  19.2M  & \checkmark &  & \checkmark & & 32.64$_{(3.59)}$ & 42.72$_{(5.01)}$ & 36.02$_{(1.09)}$ & 26.86$_{(3.51)}$ & 29.71$_{(4.78)}$ & 19.25$_{(3.50)}$    \\
\textbf{TimeLLM}  & 78M & \checkmark &  & \checkmark & & 37.43$_{(1.17)}$ & 54.93$_{(6.56)}$ & 36.59$_{(1.17)}$ & 10.18$_{(2.30)}$ & $35.21_{(6.47)}$ & 15.16$_{(2.17)}$    \\

\textbf{CAML} & 36.1M& \checkmark &  & &   \checkmark & 69.04$_{(0.18)}$ & 55.87$_{(2.72)}$ & 61.54$_{(0.30)}$ &  65.08$_{(2.56)}$ & 50.12$_{(3.05)}$ &  54.42$_{(0.94)}$ \\
\textbf{DIPOLE}  & 39M & \checkmark &  &  & \checkmark & 64.38$_{(0.89)}$ & 57.94$_{(1.15)}$ & 60.98$_{(0.27)}$ & 61.63$_{(1.03)}$ & 53.02$_{(1.18)}$ & 55.68$_{(0.49)}$   \\
\textbf{Flan-T5} &  60M   &  & \checkmark & &  \checkmark& 58.12$_{(1.11)}$  & 66.23$_{(0.72)}$ & 62.03$_{(0.54)}$ &  56.56$_{(1.03)}$ & 62.47$_{(0.76)}$   & 58.87$_{(0.71)}$  \\
\textbf{OPT} &  125M   &  & \checkmark & &  \checkmark& 42.56$_{(0.95)}$ & 69.22$_{(0.87)}$ &  52.71$_{(0.71)}$ & 47.58$_{(0.97)}$ & 65.06$_{(0.85)}$ & 52.56$_{(0.68)}$\\

\textbf{QWEN-2.5} &  500M   &  & \checkmark & &  \checkmark & 49.77$_{(1.13)}$ & 58.24$_{(0.95)}$ & 53.21$_{(0.90)}$  & 50.32$_{(1.08)}$ & 54.03$_{(0.89)}$ & 52.99$_{(0.92)}$\\
\textbf{PROMPTEHR} &75.2M &  & \checkmark &  & \checkmark & 59.29$_{(0.97)}$  & 65.53$_{(0.69)}$ & 62.24$_{(0.23)}$ & 57.44$_{(0.97)}$ & 62.87$_{(0.61)}$   & 59.10$_{(0.24)}$  \\
\textbf{LLaMA-1}  &7B & & \checkmark & \checkmark  & \checkmark & 61.42$_{(.2.08)}$  & 65.98$_{(1.53)}$  & 63.64$_{(0.41)}$  & 61.08$_{(1.54)}$  &  61.64$_{(1.27)}$  & 60.55$_{(0.44)}$    \\
\textbf{LDAM}   &41.3M& \checkmark &  & \checkmark & \checkmark & 68.00$_{(1.23)}$ & 57.12$_{(0.47)}$ & 62.18$_{(0.40)}$  & 67.38$_{(0.35)}$ & 51.50$_{(0.95)}$ & 57.44$_{(0.60)}$  \\

\textbf{FROZEN}  &265M& &  \checkmark & \checkmark & \checkmark & 61.09$_{(1.81)}$ & 64.07$_{(1.58)}$ & 62.51$_{(0.34)}$  &  59.96$_{(1.55)}$ & 59.99$_{(1.66)}$ & 
59.15$_{(0.30)}$ \\
\textbf{EHR-KnowGen} &76.9M &  & \checkmark & \checkmark & \checkmark & 60.01$_{(0.29)}$ & 65.51$_{(0.18)}$ & 62.62$_{(0.06)}$ & 58.34$_{(0.38)}$ &  61.81$_{(0.28)}$ & 59.44$_{(0.06)}$  \\
\rowcolor{gray!20} \textbf{ProMedTS} & &&&&&&&&&&\\
\rowcolor{gray!20}\textit{w/ OPT} & 316M & &  \checkmark & \checkmark & \checkmark & 54.57$_{(0.14)}$ & 54.92$_{(0.21)}$ & 54.74$_{(0.09)}$  & 55.92$_{(0.08)}$& 52.15$_{(0.13)}$ &  53.68$_{(0.07)}$ \\
\rowcolor{gray!20}\textit{w/ QWEN-2.5} & 685M& &  \checkmark & \checkmark & \checkmark & 55.37$_{(0.15)}$ & 58.97$_{(0.18)}$ &   57.82$_{(0.13)}$& 56.45$_{(0.14)}$& 56.72$_{(0.16)}$ &  55.97$_{(0.13)}$  \\

\rowcolor{gray!20}\textit{w/ Flan-T5-small} &268M& &  \checkmark & \checkmark & \checkmark & 61.32$_{(0.54)}$ & 66.65$_{(0.51)}$ & \underline{63.67}$_{(0.08)}$  & 60.35$_{(0.61)}$ & 61.62$_{(0.71)}$ & \underline{60.42}$_{(0.18)}$  \\
\rowcolor{gray!20}\textit{w/ Flan-T5-large} & 1B & &  \checkmark & \checkmark & \checkmark & 60.62$_{(0.22)}$ & 67.83$_{(0.18)}$ &  \textbf{64.02}$_{(0.11)}$ & 59.43$_{(0.37)}$ & 63.65$_{(0.54)}$ &  \textbf{60.78}$_{(0.13)}$  \\

\hline
 \multicolumn{12}{c}{\textbf{ MIMIC-IV}}  \\ 
                         \cline{1-12}

\textbf{GRU}  & 7.9M & \checkmark &  & \checkmark &  & 56.23$_{(1.13)}$ & 25.77$_{(1.58)}$ & 35.21$_{(1.36)}$ & 38.37$_{(1.90)}$ & 16.97$_{(1.22)}$ & 20.65$_{(1.32)}$  \\    
\textbf{PatchTST}  & 19.2M & \checkmark &  & \checkmark & & 27.26$_{(0.03)}$ & 57.42$_{(0.41)}$ & 36.97$_{(0.10)}$ & 20.59$_{(2.76)}$ & 43.72$_{(0.25)}$ & 21.78$_{(2.83)}$   \\
\textbf{TimeLLM}   & 78M & \checkmark &  & \checkmark & & 30.30$_{(1.78)}$ & 60.46$_{(1.98)}$ & 40.31$_{(1.20)}$ & 24.61$_{(2.21)}$ & 47.26$_{(2.37)}$ & 25.56$_{(1.60)}$    \\

\textbf{CAML}  &36.1M& \checkmark &  & &  \checkmark & 72.82$_{(0.54)}$  & 59.48$_{(0.82)}$  & 65.40$_{(0.36)}$  & 67.25$_{(0.99)}$  & 50.73$_{(1.49)}$  & 54.71$_{(1.42)}$   \\
\textbf{DIPOLE} & 39M & \checkmark &  & &  \checkmark & 72.39$_{(0.51)}$ & 61.38$_{(0.83)}$ & 66.43$_{(0.33)}$ & 70.45$_{(0.37)}$ & 55.65$_{(0.79)}$ & 60.37$_{(0.62)}$  \\

\textbf{Flan-T5}  &  60M  &  & \checkmark & &  \checkmark&  66.24$_{(0.52)}$ & 69.53$_{(0.18)}$ & 67.92$_{(0.41)}$ & 64.28$_{(0.58)}$ &  66.01$_{(0.54)}$ & 64.79$_{(0.36)}$  \\
\textbf{OPT} &  125M   &  & \checkmark & &  \checkmark& 57.02$_{(0.62)}$ & 46.53$_{(0.51)}$ &  51.25$_{(0.37)}$ &56.87$_{(0.58)}$ & 44.32$_{(0.44)}$ & 47.89$_{(0.33)}$ \\

\textbf{QWEN-2.5} &  500M   &  & \checkmark & &  \checkmark & 52.34$_{(0.62)}$ & 54.47$_{(0.51)}$ & 53.38$_{(0.46)}$  & 53.31$_{(0.55)}$ & 50.91$_{(0.43)}$ & 51.20$_{(0.38)}$\\
\textbf{PROMPTEHR} & 75.2M &  & \checkmark &   & \checkmark & 65.24$_{(0.68)}$ & 70.31$_{(0.56)}$ & 68.02$_{(0.17)}$ & 63.53$_{(0.47)}$ & 67.02$_{(0.65)}$ & 65.01$_{(0.28)}$  \\
\textbf{LLaMA-1}  &7B& &  \checkmark &\checkmark &   \checkmark &  68.54$_{(1.12)}$ & 69.54$_{(0.73)}$ & 69.29$_{(0.32)}$ &   67.53$_{(0.91)}$ & 66.24$_{(1.13)}$ & 66.21$_{(0.64)}$   \\

\textbf{LDAM}   & 41.3M& \checkmark &  & \checkmark &  \checkmark & 72.01$_{(0.85)}$ & 62.74$_{(0.62)}$ & 66.91$_{(0.20)}$ & 69.77$_{(0.18)}$ & 56.72$_{(0.69)}$ & 60.77$_{(0.48)}$   \\

\textbf{FROZEN} & 265M &  & \checkmark & \checkmark &  \checkmark &  67.81$_{(0.78)}$ & 69.08$_{(0.94)}$ & 68.42$_{(0.08)}$ & 66.27$_{(1.00)}$ & 65.21$_{(0.97)}$ & 65.30$_{(0.05)}$  \\
\textbf{EHR-KnowGen} &76.9M& &  \checkmark & \checkmark &  \checkmark & 65.80$_{(0.64)}$  & 70.85$_{(0.45)}$ &  68.16$_{(0.11)}$ & 63.82$_{(0.53)}$ & 67.24$_{(0.55)}$ & 65.11$_{(0.13)}$   \\
\rowcolor{gray!20} \textbf{ProMedTS} &&&&&&&&&&&\\
\rowcolor{gray!20}\textit{w/ OPT} & 316M  &  &  \checkmark & \checkmark & \checkmark & 54.99$_{(0.15)}$ &  55.20$_{(0.19)}$ &  55.10$_{(0.11)}$ &  56.81$_{(0.10)}$ & 52.36$_{(0.17)}$ &  53.23$_{(0.10)}$ \\

\rowcolor{gray!20}\textit{w/ QWEN-2.5} & 685M& &  \checkmark & \checkmark & \checkmark & 48.65$_{(0.25)}$ &  72.32$_{(0.17)}$ &  58.14$_{(0.16)}$  &57.35$_{(0.20)}$ &66.23$_{(0.13)}$  &  57.51$_{(0.12)}$  \\

\rowcolor{gray!20}\textit{w/ Flan-T5-small}&268M& &  \checkmark & \checkmark & \checkmark & 71.63$_{(0.46)}$ & 67.81$_{(0.85)}$ & \underline{69.69}$_{(0.18)}$ &  70.12$_{(0.47)}$ & 63.58$_{(0.79)}$ &  \underline{66.21}$_{(0.17)}$ \\
\rowcolor{gray!20}\textit{w/ Flan-T5-large}& 1B & &  \checkmark & \checkmark & \checkmark & 71.12$_{(0.31)}$ & 69.33$_{(0.42)}$ & \textbf{70.21}$_{(0.05)}$  & 70.97$_{(0.42)}$ & 65.51$_{(0.64)}$ &  \textbf{67.56}$_{(0.09)}$  \\

\toprule[1pt]
\end{tabular}

\caption{ The performance of comparative methods in the disease diagnosis tasks on MIMIC-III and MIMIC-IV. Please note CLS - classification model, GEN -generative model, Lab - lab test result, and Note - medical notes. }
\label{table1}
\end{table*}

\section{Experiments}
\subsection{Datasets and Preprocessing}
The MIMIC-III dataset \cite{johnson2016mimic} is a publicly available EHR dataset containing de-identified patients who were admitted to ICUs between 2001 and 2012. It includes medical discharge summaries, lab test results, chest x-ray images and more. Our analysis focuses on EHR data from approximately 27,000 patients, including complete medical discharge summaries and lab test results. The MIMIC-IV dataset \cite{johnson2023mimic} comprises EHR data from 2008 to 2019. We utilize approximately 29,000 EHR records from MIMIC-IV, which include complete medical discharge summaries and lab test results. 
Our study targets 25 disease phenotypes as defined in the MIMIC-III benchmark \cite{harutyunyan2019multitask}.

\paragraph{Data Pre-processing.} 
For medical notes, we extract the brief course from discharge summaries, removing numbers, noise, and stopwords. Numerical lab tests are converted into time series data using the benchmark tools \cite{Harutyunyan2019}, with missing values filled using the nearest available numbers. Time series anomaly descriptions are used with the method defined in Appendix A.2. Data splitting follows the guidelines \cite{Harutyunyan2019} using a 4:1 ratio for training and testing.

\subsection{Baseline Methods}
We benchmark our approach against a range of methods: GRU \cite{cho2014learning}, PatchTST \cite{nie2022time}, TimeLLM \cite{jintime}, CAML \cite{mullenbach2018explainable}, DIPOLE \cite{ma2017dipole}, Flan-T5 \cite{chung2024scaling}, OPT \cite{zhang2022opt}, QWEN-2.5 \cite{qwen2025qwen25technicalreport}, PROMPTEHR \cite{wang2022promptehr}, LLaMA-1-7B \cite{touvron2023llama} with anomalies input, LDAM \cite{niu2021label}, FROZEN \cite{tsimpoukelli2021multimodal}, and EHR-KnowGen \cite{niu2024ehr}. Detailed configurations of these baselines are provided in Appendix~A.3. For the disease diagnosis task, we adopt two scales of Flan-T5 \cite{chung2024scaling}, OPT-0.1B, and QWEN-2.5-0.5B as the frozen LLM to validate our model's effectiveness. To ensure a fair comparison, all baselines also employ Flan-T5-small as their backbone. Reported results are averaged over five runs with different random seeds. The statistical significance was determined at p < 0.05 by t-test. Implementation details for every model are described in Appendix~A.4, and the example of training instructions appears in Appendix~A.5. Our code is publicly available at github\footnote{\href{https://github.com/Healthcare-Data-Mining-Laboratory/PromptMedTS-V1}{https://github.com/Healthcare-Data-Mining-Laboratory/PromptMedTS-V1}}.

\begin{table*}[htbp]
\small

\setlength\tabcolsep{10.5pt}
\renewcommand\arraystretch{1}
\centering
\begin{tabular}{l|ccc|ccc}
\toprule[1pt]
\multirow{2}{*}{\textbf{Models}} & \multicolumn{3}{c|}{\textbf{Micro}} & \multicolumn{3}{c}{\textbf{Macro}}\\

            \cline{2-7}
                         & \textbf{Precision}  & \textbf{Recall}  & \textbf{F1} & \textbf{Precision}  & \textbf{Recall}  & \textbf{F1} \\ 
                          \toprule[1pt]
\multicolumn{7}{c}{\textbf{ MIMIC-III}}  \\ 
\hline

\rowcolor{gray!20} \textbf{ProMedTS } & 61.32$_{(0.54)}$ & 66.65$_{(0.51)}$ & \textbf{63.67}$_{(0.08)}$  & 60.35$_{(0.61)}$ & 61.62$_{(0.71)}$ & \textbf{60.42}$_{(0.18)}$  \\
\hline
\textbf{\textit{w/o} LAB }  & 58.91$_{(0.83)}$ & 66.59$_{(0.57)}$ & 62.34$_{(0.26)}$ & 57.32$_{(0.88)}$ & 62.56$_{(0.61)}$ & 59.05$_{(0.22)}$  \\
\textbf{\textit{w/o} ANOMALY }  & 60.09$_{(0.32)}$ &   65.03$_{(0.98)}$ & 62.44$_{(0.22)}$ & 59.13$_{(0.43)}$ &  60.46$_{(1.15)}$ & 59.11$_{(0.25)}$ \\
\hline

\multicolumn{7}{c}{\textbf{ MIMIC-IV}}  \\ 
\hline
\rowcolor{gray!20}  \textbf{ProMedTS }& 71.63$_{(0.46)}$ & 67.81$_{(0.85)}$ & \textbf{69.69}$_{(0.18)}$ &  70.12$_{(0.47)}$ & 63.58$_{(0.79)}$ &  \textbf{66.21}$_{(0.17)}$ \\
\hline
\textbf{\textit{w/o} LAB }& 67.16$_{(0.55)}$ & 69.42$_{(0.59)}$ & 68.22$_{(0.31)}$ &  65.74$_{(0.62)}$ & 64.69$_{(0.48)}$ & 64.33$_{(0.18)}$   \\
\textbf{\textit{w/o} ANOMALY} & 70.94$_{(0.37)}$ &  66.44$_{(1.37)}$ &   68.47$_{(0.12)}$ & 68.95$_{(0.79)}$    & 62.45$_{(0.96)}$ &  65.13$_{(0.12)}$  \\

\toprule[1pt]
\end{tabular}

\caption{Ablation studies on different modality input and alignment designs for disease diagnosis.}
\label{table2}
\end{table*}

\begin{table*}[t]
\small
\setlength\tabcolsep{10.5pt}
\renewcommand\arraystretch{1}
\centering

\begin{tabular}{l|ccc|ccc}
\toprule[1pt]
\multirow{2}{*}{\textbf{Models}} &\multicolumn{3}{c|}{\textbf{Micro}}  &\multicolumn{3}{c}{\textbf{Macro}} \\

         \cline{2-7}
                         & \textbf{Precision}  & \textbf{Recall}  & \textbf{F1} & \textbf{Precision}  & \textbf{Recall}  & \textbf{F1} \\ 
\toprule[1pt]

\multicolumn{7}{c}{\textbf{ MIMIC-III}}  \\ 
\hline
\rowcolor{gray!20}  \textbf{ProMedTS} & 61.32$_{(0.54)}$ & 66.65$_{(0.51)}$ & \textbf{63.67}$_{(0.08)}$  & 60.35$_{(0.61)}$ & 61.62$_{(0.71)}$ & \textbf{60.42}$_{(0.18)}$  \\
\hline
\textbf{\textit{w/o} CONTRAST}  & 60.24$_{(0.25)}$ & 66.00$_{(0.39)}$ & 62.99$_{(0.07)}$  & 59.92$_{(0.60)}$ & 61.41$_{(0.51)}$ & 59.73$_{(0.07)}$  \\
\textbf{\textit{w/o} MATCH}  &  60.12$_{(0.58)}$ &  66.14$_{(1.50)}$ & 62.96$_{(0.02)}$  & 59.70$_{(1.18)}$ & 61.37$_{(1.34)}$ & 59.65$_{(0.11)}$  \\
\textbf{\textit{w/o} GEN}& 59.95$_{(0.38)}$ &  66.15$_{(0.27)}$ & 62.89$_{(0.19)}$  &   59.57$_{(0.55)}$ &  61.32$_{(0.30)}$ &  59.61$_{(0.20)}$ \\
\hline

\multicolumn{7}{c}{\textbf{ MIMIC-IV}}  \\ 
\hline
\rowcolor{gray!20}  \textbf{ProMedTS }& 71.63$_{(0.46)}$ & 67.81$_{(0.85)}$ & \textbf{69.69}$_{(0.18)}$ &  70.12$_{(0.47)}$ & 63.58$_{(0.79)}$ &  \textbf{66.21}$_{(0.17)}$ \\
\hline
\textbf{\textit{w/o} CONTRAST} &  70.19$_{(0.25)}$ &  66.22$_{(0.39)}$ & 68.61$_{(0.09)}$ &  69.05$_{(0.24)}$ & 62.40$_{(0.35}$ & 65.21$_{(0.12)}$\\
\textbf{\textit{w/o} MATCH}& 70.79$_{(0.34)}$ & 66.49$_{(0.38)}$ & 68.67$_{(0.15)}$ & 68.91$_{(0.73)}$ & 62.25$_{(0.48)}$ & 65.47$_{(0.15)}$\\
\textbf{\textit{w/o} GEN}  &  71.30$_{(0.29)}$ &  65.79$_{(0.57)}$ &  68.44$_{(0.13)}$ & 69.14$_{(0.48)}$ & 62.05$_{(0.54)}$ &  65.03$_{(0.13)}$\\

\toprule[1pt]
\end{tabular}
\caption{Ablation studies on the effectiveness of different loss functions of our model for disease diagnosis. }
\label{table3}
\end{table*}
\subsection{Disease Diagnosis Performance}
Table~\ref{table1} shows the evaluation performance of disease diagnosis for our model ProMedTS with all baselines.  First, for a single modality model, TimeLLM performs strongly with lab tests, highlighting the value of time-series inputs for LLMs in disease diagnosis. Text-based methods (e.g., Flan-T5) generally outperform time-series approaches, suggesting that medical notes capture richer disease-related information. Multimodal models (e.g., EHR-KnowGen, LLaMA) exceed single-modality baselines (e.g., TimeLLM, PROMPTEHR), confirming the benefits of integrating text and time series. Generative approaches (e.g., TimeLLM, LLaMA, EHR-KnowGen) also outperform classification-based methods. Although LLaMA performs well, its higher variance and parameter requirements reduce practicality.

Notably, the performance of our ProMedTS model, which employs Flan-T5-small as its backbone LLM, exceeds that of all baseline methods (especially LLaMA-1-7B) across both micro and macro F1 scores. This outcome emphasizes the efficiency and effectiveness of our approach in enabling LLMs to achieve multimodal understanding for disease diagnosis tasks. Additionally, ProMedTS consistently improves disease diagnosis performance across various backbone LLMs (Flan-T5-small\&large, OPT, QWEN), achieving an average F1 score improvement of around 3\%. These findings underscore our model's scalability, robustness, and the generalizability of our approach.

\subsection{Ablation Studies}

\subsubsection{Effect of Modality Alignment in ProMedTS}

This section presents ablation studies to evaluate each module in ProMedTS. ProMedTS \textit{w/o} LAB excludes lab test, removing modality alignment with anomaly descriptions and medical notes. ProMedTS \textit{w/o} ANOMALY removes alignment with anomaly descriptions while keeping alignment between lab test and medical notes to assess the impact of self-supervision. Table~\ref{table2} summarizes the results, showing that ProMedTS \textit{w/o} LAB  suffers a significant drop in F1 scores, highlighting the importance of the lab test. ProMedTS \textit{w/o} ANOMALY also shows reduced performance, highlighting the challenges of aligning modalities from discrete and continuous encoding spaces and the adverse effects of misalignment on multimodal understanding. These findings emphasize the crucial role of both lab tests and anomaly captions as input for learning accurate time series embeddings, which in turn assist LLMs in disease diagnosis.

\subsubsection{Impact of Self-Supervised Loss Functions}

Table~\ref{table3} summarizes an ablation study on the loss functions in ProMedTS. Both ProMedTS \textit{w/o} CONTRAST and ProMedTS \textit{w/o} MATCH show slight declines in F1 scores, emphasizing the importance of $\mathcal{L}_{contrast}$ for aligning and unifying time series and textual inputs within a shared latent space. The results also underscore the role of $\mathcal{L}_{match}$ in intra-modal alignment, ensuring the distinctiveness of time series data by aligning lab tests with time series prompt embeddings. Notably, ProMedTS \textit{w/o} GEN exhibits a significant drop in F1 scores, highlighting the critical role of $\mathcal{L}_{gen}$ in refining prompt embeddings and integrating temporal information from time series data and anomaly descriptions.

\begin{figure}[htbp]
\hspace{-0.5cm}
\centerline{\includegraphics[scale=0.4]{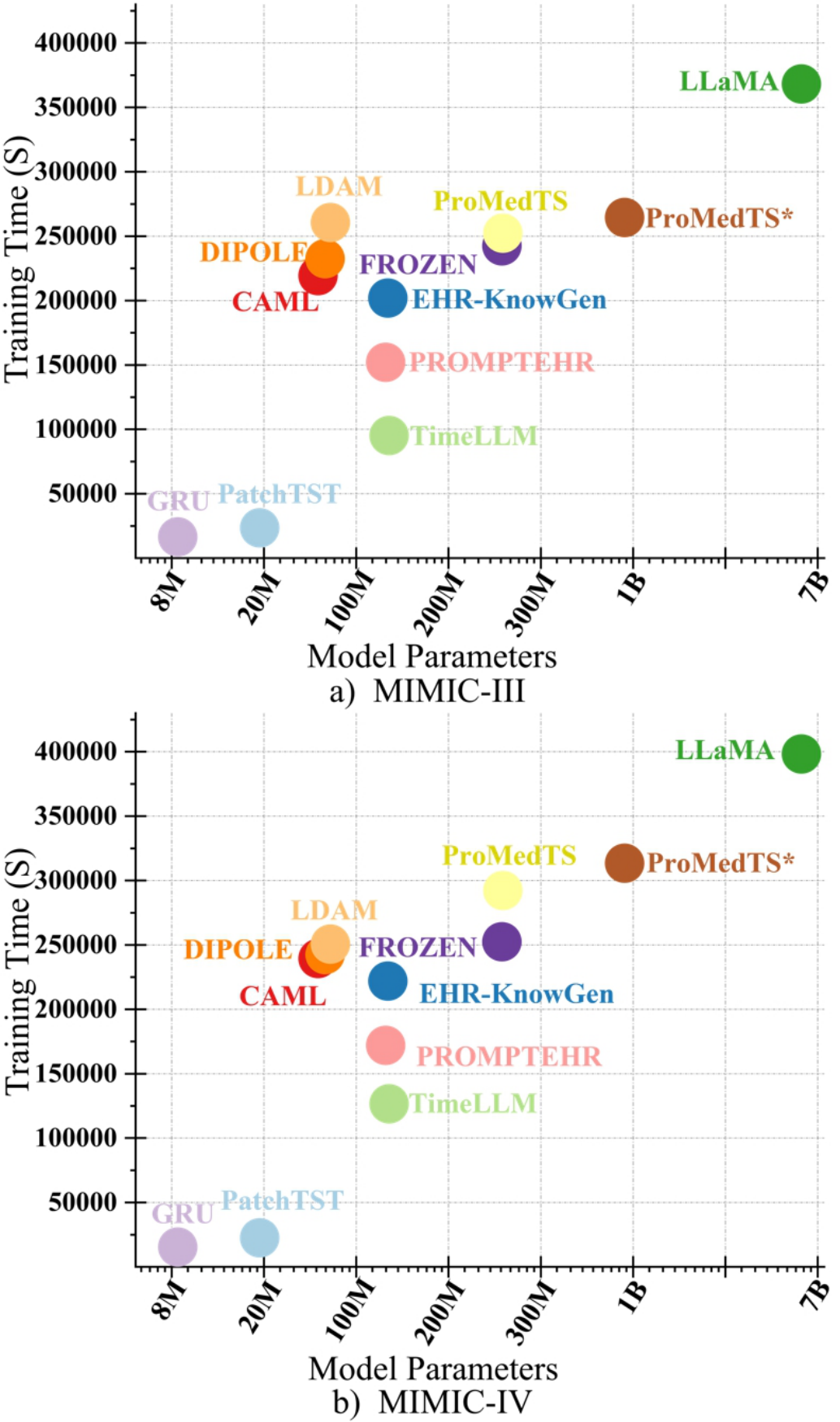}}
\caption{ The model parameters and computation time of all baselines.}
\label{complexity}
\end{figure}

\subsection{Model Efficiency and Complexity} \label{complexity analysis}
Figure~\ref{complexity} illustrates the parameter counts and computation times of baseline models on the two datasets. Our model, ProMedTS, matches the parameter counts and computation times of multimodal baselines such as LDAM and FROZEN, while using 25× fewer parameters and requiring one-third less training time than LLaMA, all while achieving superior diagnostic performance, highlighting its efficiency and effectiveness in language-time series multimodal alignment and fusion.

In addition, we conducted an efficiency evaluation using EHR-KnowGen with different backbone LLMs. As shown in Table~\ref{table_efficiency}, ProMedTS outperformed EHR-KnowGen by an average of 2\% improvement in F1 scores while maintaining comparable training times with backbone LLMs of OPT and QWEN-2.5. This result supports the effectiveness and efficiency of our model in disease diagnosis.

 \subsection{Sensitivity Analysis of Time Series Prompt Length}
We performed a sensitivity analysis to examine the impact of the time series prompt embedding length ($N_p$) on the performance of ProMedTS in disease diagnosis. Table~\ref{sensitivity} shows the F1 scores for embedding lengths of 12, 24, and 36. Slight fluctuations are observed in both micro and macro F1 scores across datasets. The optimal embedding length is 24 for both datasets, consistent with the configuration used in our experiments.

\begin{table}[t]
\small
\setlength\tabcolsep{5pt}
\renewcommand\arraystretch{1}
\centering

\begin{tabular}{l|c|c|c}
\toprule[1pt]

\multirow{2}{*}{\textbf{Models}} & \multirow{1}{*}{\textbf{Training}}  &\multirow{2}{*}{\textbf{Micro F1}} &\multirow{2}{*}{\textbf{Macro F1}}  \\
& \textbf{Time (h)}  & & \\
  \hline    
 \multicolumn{4}{c}{\textbf{ MIMIC-III}}  \\ 
                         \cline{1-4}

\textbf{EHR-KnowGen} &&&\\
\textit{w/ OPT} & 8h & 53.11 &  52.85 \\
\textit{w/ QWEN-2.5} & 12h &  54.09  & 51.45  \\
\rowcolor{gray!20} \textbf{ProMedTS} &&&\\
\rowcolor{gray!20}\textit{w/ OPT} & 8h  &  54.74 & 53.68  \\
\rowcolor{gray!20}\textit{w/ QWEN-2.5} & 14h &   57.82 &  55.97 \\

\hline
 \multicolumn{4}{c}{\textbf{ MIMIC-III}}  \\ 
                         \cline{1-4}

\textbf{EHR-KnowGen} &&&\\
\textit{w/ OPT} & 8.5h  & 52.46 &  50.19 \\
\textit{w/ QWEN-2.5} & 13h &   54.19	& 52.77 \\
\rowcolor{gray!20} \textbf{ProMedTS} &&&\\
\rowcolor{gray!20}\textit{w/ OPT} & 9h  & 55.10	& 53.23  \\
\rowcolor{gray!20}\textit{w/ QWEN-2.5} & 14.5h &  58.14 &	57.51 \\

\toprule[1pt]
\end{tabular}

\caption{ Performance and Efficiency Comparison on Disease Diagnosis Tasks with Different LLMs. }
\label{table_efficiency}
\end{table}

\begin{table}[t]
\small
\setlength\tabcolsep{18.5pt}
\renewcommand\arraystretch{1}
\centering

\begin{tabular}{c|c|c}
\toprule[1pt]
 \multicolumn{1}{c|}{$N_p$} & \multicolumn{1}{c}{\textbf{Micro F1}} & \multicolumn{1}{|c}{\textbf{Macro F1}}\\
                          \toprule[1pt]
\multicolumn{3}{c}{\textbf{ MIMIC-III}}  \\ 
\hline
 12 & 63.09$_{(0.06)}$ & 59.69$_{(0.12)}$  \\
 \rowcolor{gray!20} 24 & 63.67$_{(0.08)}$ & 60.42$_{(0.18)}$ \\
 36 & 63.32$_{(0.09)}$ &  59.96$_{(0.15)}$ \\
\hline
\multicolumn{3}{c}{\textbf{ MIMIC-IV}}  \\ 
\hline
 12 & 68.98$_{(0.15)}$ & 65.43$_{(0.18)}$  \\
\rowcolor{gray!20} 24 & 69.69$_{(0.18)}$ & 66.21$_{(0.17)}$ \\
 36 & 69.41$_{(0.19)}$ & 65.91$_{(0.20)}$ \\
\toprule[1pt]
\end{tabular}
\caption{Sensitivity analysis on different lengths of time series prompt embedding. }
\label{sensitivity}
\end{table}

\subsection{Evaluating the Role of Anomaly Descriptions}
To highlight the advantages of using lab test anomaly captions over raw numerical time series values in LLMs, we evaluate Flan-T5-small with both input types. Table~\ref{tsvsanomaly} presents the evaluation results on the MIMIC-III and MIMIC-IV datasets for disease diagnosis. The results show that Flan-T5 achieves over a 2\% improvement in Micro F1 score when using anomaly captions, demonstrating that LLMs interpret anomaly captions more effectively than raw numerical values in time series lab test data. Additionally, the inclusion of time series prompts underscores the effectiveness of our model, ProMedTS, in capturing both fine-grained and coarse-grained temporal information from lab test results for disease diagnosis.

\begin{table}[t]
\small
\setlength\tabcolsep{8pt}
\renewcommand\arraystretch{1}
\centering

\begin{tabular}{c|c|c}
\toprule[1pt]
 \multicolumn{1}{c|}{Lab Test Input} & \multicolumn{1}{c}{\textbf{Micro F1}} & \multicolumn{1}{|c}{\textbf{Macro F1}}\\
                          \toprule[1pt]
\multicolumn{3}{c}{\textbf{ MIMIC-III}}  \\ 
\hline
Numerical Values & 32.21$_{(1.33)}$ & 23.53$_{(1.21)}$  \\
Anomaly captions & 35.19$_{(0.92)}$ & 24.75$_{(0.76)}$ \\
Time series prompts & 36.11$_{(1.14)}$ & 25.47$_{(1.02)}$ \\
 \hline
\multicolumn{3}{c}{\textbf{ MIMIC-IV}}  \\ 
\hline
Numerical Values & 37.75$_{(1.46)}$ & 26.10$_{(1.09)}$  \\
Anomaly captions & 39.56$_{(1.14)}$ & 27.22$_{(0.77)}$ \\
Time series prompts & 40.14$_{(1.05)}$ & 28.43$_{(0.91)}$ \\
\toprule[1pt]
\end{tabular}
\caption{Micro and Macro F1 Scores Across Various Lab Test Input Types on the LLM for Disease Diagnosis}
\label{tsvsanomaly}
\end{table}
\section{Conclusion and Future Work}

In this paper, we introduce ProMedTS, a lightweight and effective modality fusion framework that leverages self-supervised time series prompt learning for multimodal EHR integration. By bridging the modality gap between medical notes and lab test results, ProMedTS enables LLMs to process structured and unstructured medical data more effectively. Its three key modules and loss functions advance language–time series integration in healthcare, providing a scalable and adaptable approach for real-world clinical applications. Evaluation on two real-world EHR datasets demonstrates that ProMedTS significantly outperforms existing models in disease diagnosis, underscoring its potential to enhance LLMs' clinical decision-making and improve patient care.  In future work, we plan to extend our approach to apply to larger and more diverse datasets, such as Time-MMD \cite{liu2024time}, conduct experiments with different clinical tasks, explore additional LLM architectures, and investigate further improvements in modality alignment techniques.

\section*{Limitations}

While this study focuses on modality alignments and their application in downstream tasks, enhancing the explainability of disease diagnosis remains an area for future work, where we plan to incorporate the Chain-of-Thought rationale \cite{wei2022chain} via knowledge distillation \cite{hsieh2023distilling,lin2023beneath}. Furthermore, our study primarily targets higher-level disease phenotypes \cite{slee1978international}, which could be expanded to more downstream tasks.

\section*{Ethics Statement}

\paragraph{Data Privacy:} While the datasets utilized in our research, such as MIMIC-III and MIMIC-IV, are publicly accessible and feature de-identified patient data, accessing these datasets still requires passing the \href{https://about.citiprogram.org/}{CITI examination} and applying for the data through \href{https://physionet.org/}{PhysioNet}.

\section*{Acknowledgments}
This work is supported by Tencent Rhino-Bird Focused Research Program (Value-aligned Credible Large Language Model) and RMGS project (Artificial Intelligence and Big Data Analytics for Social Good), and is also supported by the National Natural Science Foundation of China (No.62432006, No.62276159).

\bibliography{custom}
\bibliographystyle{acl_natbib}
\appendix

\section{Appendix}
\label{sec:appendix}

\subsection{Algorithm}
The training procedure to optimize ProMedTS by minimizing the loss defined in Equation \eqref{eq10} is shown in Algorithm 1.
\begin{algorithm}[htbp]
    \caption{The ProMedTS Model}
    \label{alg:algorithm}
    \begin{algorithmic}[1]
        \STATE {\textbf{Input}: Given lab test $\bm{X}$ and medical note $\bm{M}$ denote the EHRs input. $\bm{P}$ consists of a set of learnable prompt embeddings.}
        \WHILE{not converge}
            \FOR{mini-batch $B$}
                \STATE {Obtain the time series anomaly caption $\bm{\mathcal{T}}$ using equation \eqref{eq1}.}
                \STATE {Obtain multimodal textual embedding $\bm{E}_f$ using equations \eqref{eq2} and \eqref{eq3}.}
                \STATE {Calculate the contrastive loss $\mathcal{L}_{contrast}$ between lab test, anomalies, and medical notes using equations \eqref{eq4}, \eqref{eq5}, and \eqref{eq6}.}
                \STATE {Calculate the matching loss $\mathcal{L}_{match}$ between lab test and anomalies using equations \eqref{eq7} and \eqref{eq8}.}
                \STATE {Calculate the generation loss $\mathcal{L}_{gen}$ between lab test and anomalies using equation \eqref{eq9}.}
            \ENDFOR
            \STATE{Update parameters by minimizing the total loss $\mathcal{L}_{total}$ defined in Equation \eqref{eq10} by  using the AdamW optimizer \cite{loshchilov2018decoupled} for patients in each batch.}
        \ENDWHILE
    \end{algorithmic}
\end{algorithm}

\subsection{Lab Test Anomaly Caption}
Time series anomaly descriptions are generated using the IQR method \cite{vinutha2018detection} to identify anomalies, capturing their timing and polarity (above or below standard values) and describing them with handcrafted templates. To caption the lab test anomaly in textual format, we design several text templates to describe the lab test anomalies. All templates are illustrated in Table~\ref{anomaly}.
\begin{table*}[htbp]
\setlength\tabcolsep{2.2pt}
\renewcommand\arraystretch{1}
\centering
\begin{tabular}{l}
\toprule[1pt]
If lab test value is not an abnormal value:  \\
\textit{\{Lab features\} is normal all the time.}\\
\hline
If the lab test value is an abnormal value higher than the standard: \\
\textit{\{Lab features\} is higher than normal \{number of times\} times.}\\
\hline
If the lab test value is an abnormal value lower than the standard: \\
\textit{\{Lab features\} is lower than normal \{number of times\} times.}\\
\hline
If the lab test value is an abnormal that include both higher and lower than the standard value: \\
\textit{\{Lab features\} is higher than normal \{number of times\} times and lower than normal}\{number  of \\ \textit{ times\} times.}\\
\hline

\toprule[1pt]
\end{tabular}
\caption{Lab test anomaly caption template. }

\label{anomaly}
\end{table*}

\begin{table*}[htbp]
\setlength\tabcolsep{1.7pt}
\renewcommand\arraystretch{1}
\centering
\begin{tabular}{l}
\toprule[1pt]
\textbf{Diagnose disease from the following medical notes and lab test:}\\
\textbf{Medical Notes:} \\

Diagnose disease from the following medical notes: woman decompensated etoh cirrhosis \\ initially   fatigue ascites hydrothorax uti patient complicated hospital transferred icu hypoxemic \\ respiratory failure  developed hypotension setting bleeding esophageal varices underwent tips \\ banding episodes gi bleeding … prevention outpatient electrolyte abnormalities refeeding syndrome…  \\
contact narrow urine culture positive plan continue day add fosfomycin …\\
\textbf{Lab test Anomaly Descriptions }: $\leftarrow$ {\texttt{Only used during multimodal alignment}}\\

diastolic blood pressure is higher than normal one times, fraction inspired oxygen is higher than \\
normal forty-six times, glucose is higher than normal one times, heart rate is normal all the time, \\
mean blood pressure is normal all the time, oxygen saturation is normal all the time, respiratory rate \\
is higher than normal four times, systolic blood pressure is normal all the time, temperature is normal \\
all the time, weight is normal all the time, ph is normal all the time \\

\textbf{Lab Test:} \\
\texttt{[[73.           1.         120.          87.          73.        94.          24.         105.          37.          81.         7.53   ],} \\
\texttt{[ 79.           1.         120.          87.          94.        94.          30.         131.          37.          81.         7.53     ],} \\
\texttt{[73.           1.         217.          87.          73.        94.          24.         105.          37.           81.         7.53     ],}\\
        \texttt{ ..., } \\
\texttt{[73.           1.         120.          87.          73.       94.          22.         105.          37.            81.        7.53      ]]}\\ 
\hline
\textbf{Diagnosis:} \\
Diagnosed Results: Acute and unspecified renal failure, Fluid and electrolyte disorders, \\Septicemia (except in labor),  Shock, Chronic obstructive pulmonary disease and bronchiectasis, \\ Disorders of lipid metabolism, 
 Cardiac dysrhythmias, Congestive heart failure; nonhypertensive,\\
 Diabetes mellitus with complications, Other liver diseases.\\
\toprule[1pt]
\end{tabular}
\caption{The Example of Training Instruction Template}
\label{instruction}
\end{table*}

\begin{figure*}[htbp]
\hspace{-0.5cm}
\centerline{\includegraphics[scale=0.54]{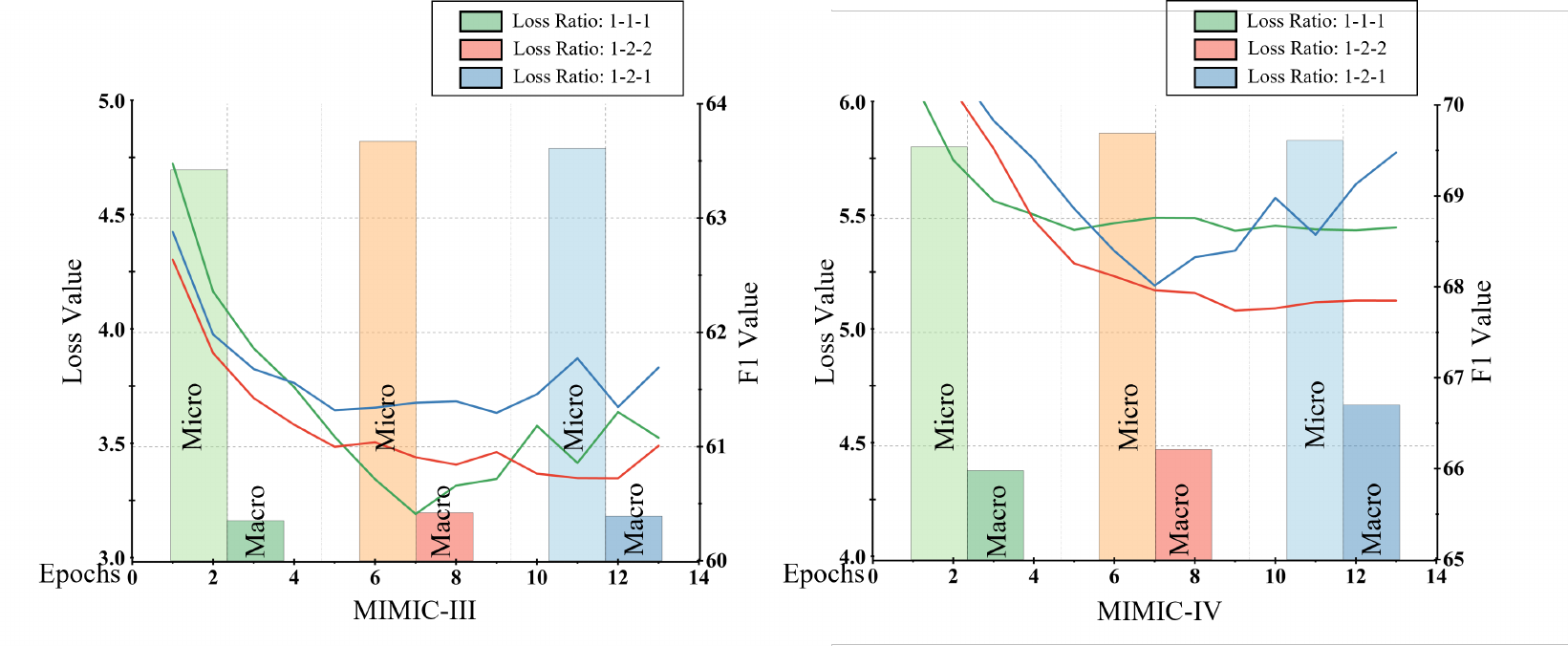}}
\caption{Sensitivity analysis of varying ratios in loss function components, showing Micro and Macro F1 scores for downstream tasks.}
\label{loss_sensivity}
\end{figure*}

\subsection{Baseline Models}

\begin{itemize}

\item \textbf{GRU}: The Gated Recurrent Unit (GRU) \cite{cho2014learning}, a variant of recurrent neural networks (RNNs), employs two gates to capture both long-term and short-term temporal features effectively.

\item \textbf{PatchTST}: PatchTST \cite{nie2022time} is a transformer-based time series encoder designed for long-term forecasting. It segments time series into subseries-level patches, treating each as a token within the transformer architecture.

\item \textbf{TimeLLM}: TimeLLM \cite{jintime} is an LLM-based time series prediction model that reprograms input time series into text prototypes before processing them with a frozen LLM. It achieves state-of-the-art performance in mainstream forecasting tasks, particularly in few-shot and zero-shot scenarios.

\item \textbf{CAML}: The Convolutional Attention for Multi-Label classification (CAML) \cite{mullenbach2018explainable} is a classical model for classifying medical notes, incorporating a cross-attention mechanism and label embeddings to enhance interpretability. For a fair comparison with more recent language models, its original embedding layer is replaced with one from T5.

\item \textbf{DIPOLE}: DIPOLE \cite{ma2017dipole} is a classic disease prediction model that utilizes two Bi-directional RNNs. It incorporates an attention mechanism to integrate information from both past and future hospital visits. For a fair comparison with more recent language models, its original embedding layer has been replaced with one from T5.

\item \textbf{Flan-T5}: showcased within the scaling instruction-fine-tuning framework for language models \cite{chung2024scaling}. It benefits from training on a wide array of datasets geared toward tasks like summarization and question answering.

\item \textbf{OPT}:OPT is a decoder-only model, designed to match the performance of GPT-3 \cite{brown2020language} while requiring only 1/7 of its computational resources. Trained primarily on around 180 billion tokens, OPT excels in tasks like poetry generation, dialogue, few-shot translation, and programming \cite{zhang2022opt}.

\item \textbf{QWEN-2.5}:  QWEN-2.5 is a cutting-edge language model, expanding its pre-training dataset to 18 trillion tokens to enhance common sense, expert knowledge, and reasoning. It undergoes extensive supervised fine-tuning and multi-stage reinforcement learning, achieving top-tier performance in benchmarks for language understanding, reasoning, mathematics, coding, and human preference alignment \cite{qwen2025qwen25technicalreport}.

\item \textbf{PROMPTEHR}: PROMPTEHR \cite{wang2022promptehr} introduces a novel approach in generative models for electronic health records (EHRs), implementing conditional prompt learning. In this study, the model is specifically geared towards disease diagnosis.

\item \textbf{LLaMA}: LLaMA-7B \cite{touvron2023llama}, one of the leading large language models, is enhanced by Reinforcement Learning with Human Feedback (RLHF) and instructive tuning. It is fine-tuned for disease diagnosis in this study, demonstrating its versatility in various NLP tasks.

\item \textbf{LDAM}: LDAM \cite{niu2021label} leverages multimodal inputs, combining laboratory testing results and medical notes for disease risk prediction. It utilizes label embedding to effectively integrate these two modalities.

\item \textbf{FROZEN}: FROZEN \cite{tsimpoukelli2021multimodal} represents the cutting-edge multimodal vision-language models for few-shot learning. In our study, it is adapted to the disease diagnosis task using inputs from lab test results and medical notes.

\item \textbf{EHR-KnowGen}: EHR-KnowGen \cite{niu2024ehr}, touted as the state-of-the-art in EHR multimodal learning models, focuses on disease diagnosis generation. For this study, external domain knowledge is excluded to ensure a fair comparison.
\end{itemize}

\subsection{Implementation Details}
In experiments, we utilized the PyTorch framework version 2.0.1, operating on a CUDA 11.7 environment. We employed the AdamW optimizer with a starting learning rate of $1e^{-5}$ and a weight decay parameter of 0.05. Additionally, we implemented a warm-up strategy covering 10\% of the training duration. Our experiments were conducted on high-performance NVIDIA Tesla V100 GPUs. Within the ProMedTS model, we used 24 time series prompt embeddings, each with a dimensionality of 768. The model's hidden layer size was maintained at 768 for modality alignment and adjusted to 512 for downstream tasks. To standardize the time series data input, we padded all lab test results to a uniform length of 1000 time steps, allowing us to divide the data into 125 patches, with each patch containing 8 time steps. All LLMs are fine-tuned on two MIMIC datasets \cite{johnson2016mimic,johnson2023mimic} and then frozen for downstream tasks. 

\subsection{Training Instruction Template}
Table~\ref{instruction} illustrates the training instruction template for our model ProMedTS for disease diagnosis on MIMIC-III and MIMIC-IV datasets.

\subsection{Sensitivity Analysis of Varying Ratios in Loss Function Components}

To examine the impact of different combinations of the three loss functions, $\mathcal{L}_{contrast}$, $\mathcal{L}_{match}$, and $\mathcal{L}_{gen}$, on the downstream performance, we perform a sensitivity analysis using three sets of loss ratios: 1:1:1, 1:2:2, and 1:2:1 on MIMIC-III and MIMIC-IV datasets. Since the value of $\mathcal{L}_{contrast}$ is typically larger than those of $\mathcal{L}_{match}$ and $\mathcal{L}_{gen}$, we assign greater weights to $\mathcal{L}_{match}$ and $\mathcal{L}_{gen}$. Figure~\ref{loss_sensivity} presents the results, where lines indicate the variation in the sum of the three loss functions on the testing dataset and bars represent the Micro and Macro F1 scores. The figure reveals that varying the weight ratios of the three loss functions has minimal impact on model convergence and the performance of downstream disease diagnosis tasks.
\end{document}